\documentclass{article}

\usepackage{amsmath}
\usepackage[utf8]{inputenc} 
\usepackage[T2A,T1]{fontenc}  
\usepackage{lmodern}
\usepackage{hyperref}       
\usepackage{url}            
\usepackage{booktabs}       
\usepackage{amsfonts}       
\usepackage{nicefrac}       
\usepackage{microtype}      
\usepackage{graphicx} 
\usepackage{xcolor}
\usepackage{subfigure}
\usepackage{mathrsfs}
\usepackage{bm}
\usepackage[margin=1.0in]{geometry}
\usepackage{color}
\usepackage{CJKutf8}
\usepackage[russian,english]{babel}
\newcommand{\todo}[1]{}
\renewcommand{\todo}[1]{{\color{red} TODO: {#1}}}

\usepackage{amsthm}
\usepackage{amsmath}
\usepackage{amssymb}
\usepackage{color}
\usepackage{pgfplots, pgfplotstable}

\newcommand{\comment}[1]{}

\def\int{\mathrm{int}}

\title{Google's Multilingual Neural Machine Translation System: Enabling Zero-Shot Translation}

\author{
  Melvin Johnson, Mike Schuster, Quoc V. Le, Maxim Krikun, Yonghui Wu,\\
  Zhifeng Chen, Nikhil Thorat\\
  \texttt{melvinp,schuster,qvl,krikun,yonghui,zhifengc,nsthorat@google.com} \and
  Fernanda Vi\'egas,
  Martin Wattenberg,
  Greg Corrado,\\
  Macduff Hughes,
  Jeffrey Dean}

\date{}
\begin{document}
\maketitle

\begin{abstract}
We propose a simple solution to use a single Neural Machine
Translation (NMT) model to translate between multiple languages.
Our solution requires no changes
to the model architecture from a standard NMT system but instead introduces an
artificial token
at the beginning of the input sentence to specify the required target language.
The rest of the model, which
includes an encoder, decoder and attention module, remains unchanged and is
shared across
all languages. Using a shared wordpiece vocabulary, our approach enables
Multilingual NMT using a single model without any increase in parameters,
which is significantly simpler than previous proposals for Multilingual NMT.
On the WMT'14 benchmarks, a
single multilingual model achieves comparable performance for
English$\rightarrow$French and surpasses state-of-the-art results for
English$\rightarrow$German. Similarly, a single multilingual model surpasses
state-of-the-art results for French$\rightarrow$English and
German$\rightarrow$English on WMT'14 and WMT'15 benchmarks, respectively.
On production corpora, multilingual models of up
to twelve language pairs allow for better translation of many individual
pairs. In addition to improving the translation quality of language pairs
that the model was trained with, our models can also learn
to perform implicit bridging between language pairs never seen explicitly
during training, showing that transfer learning and zero-shot translation
is possible for neural translation. Finally, we show analyses that hints
at a universal interlingua representation in our models and show some
interesting examples when mixing languages.

\end{abstract}
\section{Introduction}
\label{intro}
End-to-end Neural Machine Translation
(NMT)~\cite{sutskever2014sequence,BahdanauCB15,cho2014learning}
is an approach to machine translation that has rapidly gained adoption in
many large-scale settings~\cite{DBLP:journals/corr/ZhouCWLX16,wu2016,systran_paper}. Almost all such systems are built
for a single language pair --- so far there has not been a sufficiently simple
and efficient way to handle multiple language pairs using a single model without
making significant changes to the basic NMT architecture.


In this paper we introduce a simple method to translate between multiple
languages using a single model, taking advantage of multilingual data to
improve NMT for all languages involved. Our method requires no change to the
traditional NMT model architecture. Instead, we add an artificial token to
the input sequence to indicate the required target language, a simple
amendment to the data only. All other parts of the system --- encoder, decoder,
attention, and shared wordpiece vocabulary as described in~\cite{wu2016} ---
stay exactly the same. This method has several attractive benefits:
\begin{itemize}
\item {\bf Simplicity}:
Since no changes are made to the architecture of
the model, scaling to more languages is trivial --- any new data is simply
added, possibly with over- or under-sampling such that all languages are
appropriately represented, and used with a new token if the target language
changes. Since no changes are made to the training procedure, the mini-batches
for training are just sampled from the overall mixed-language training data
just like for the single-language case. Since no a-priori decisions about how
to allocate parameters for different languages are made the system adapts
automatically to use the total number of parameters efficiently to minimize the
global loss. A multilingual model architecture of this type also simplifies
production deployment significantly since it can
cut down the total number of models necessary when dealing with multiple
languages. Note that at Google, we support a total of over 100 languages as
source and target, so theoretically $100^2$ models would be necessary for the
best possible translations between all pairs, if each model could only support
a single language pair. Clearly this would be problematic in a production
environment. Even when limiting to translating to/from English only,
we still need over 200 models.
Finally, batching together many requests from potentially
different source and target languages can significantly improve efficiency of
the serving system. In comparison, an alternative system that requires
language-dependent encoders, decoders or attention modules does not have any
of the above advantages.
\item {\bf Low-resource language improvements}: In a multilingual NMT model,
all parameters are implicitly shared by all the language pairs being
modeled. This forces the model to generalize across language boundaries
during training. It is observed that when language pairs
with little available data and language pairs with abundant
data are mixed into a single model, translation quality on the low
resource language pair is significantly improved.
\item {\bf Zero-shot translation}: A surprising benefit of modeling several
language pairs in a single model is that the model can learn to
translate between language pairs it has never seen in this combination during
training (zero-shot translation) --- a working
example of transfer learning within neural translation models. For example,
a multilingual NMT model trained with Portuguese$\rightarrow$English and
English$\rightarrow$Spanish examples can generate reasonable
translations for Portuguese$\rightarrow$Spanish although it has not seen any
data for that language pair. We show that the quality of zero-shot
language pairs can easily be improved with little additional data of the
language pair in question (a fact that has been previously confirmed for a
related approach which is discussed in more detail in the next section).
\end{itemize}

In the remaining sections of this paper we first discuss related work and
explain our multilingual system architecture in more detail. Then, we go through
the different ways of merging languages
on the source and target side in increasing difficulty (many-to-one,
one-to-many, many-to-many), and
discuss the results of a number of experiments on WMT benchmarks, as well
as on some of Google's large-scale production datasets. We present results from
transfer learning experiments and show how implicitly-learned bridging
(zero-shot translation) performs in comparison to explicit bridging (i.e.,
first translating to a common language like English and then translating from
that common language into the desired target language) as typically used in
machine translation systems. We describe visualizations of the new system
in action, which provide early evidence of shared semantic representations (interlingua)
between languages. Finally we also show some interesting applications of mixing
languages with examples: code-switching on the source side and weighted target
language mixing, and suggest possible avenues for further exploration.

\section{Related Work}
\label{relwork}
Interlingual translation is a classic method in machine
translation~\cite{richens1958interlingual,hutchins1992introduction}. Despite
its distinguished history, most practical applications of machine
translation have focused on individual language pairs because it was simply
too difficult to build a single system that translates reliably
from and to several languages.

Neural Machine Translation (NMT) \cite{kalchbrenner2013recurrent} was shown to be a
promising end-to-end learning approach in
\cite{sutskever2014sequence,BahdanauCB15,cho2014learning} and was
quickly extended to multilingual machine translation in various ways.

An early attempt is the work in~\cite{dong2015multi}, where the authors modify
an attention-based encoder-decoder approach to perform multilingual NMT by
adding a separate decoder and attention mechanism for each target language.
In~\cite{luong2015multi} multilingual training in a multitask learning setting
is described. This model is also an encoder-decoder network, in this case
without an attention mechanism. To make proper use 
of multilingual data, they extend their model with multiple
encoders and decoders, one for each supported source and target language.
In~\cite{caglayan-EtAl:2016:WMT} the authors incorporate multiple modalities
other than text into the encoder-decoder framework.

Several other approaches have been proposed for multilingual training,
especially for
low-resource language pairs. For instance, in~\cite{ZophK16} a form of
multi-source translation was proposed where the model has multiple different
encoders and different attention mechanisms for each source
language. However, this work requires the presence of a multi-way
parallel corpus between all the languages involved, which is
difficult to obtain in practice. Most closely
related to our approach is~\cite{FiratCB16} in which the authors
propose multi-way multilingual NMT using a single shared attention
mechanism but multiple encoders/decoders for each source/target language.
Recently in~\cite{lee2016characternmt} a CNN-based character-level encoder
was proposed which is shared across multiple source languages. However, this
approach can only perform translations into a single target language.

Our approach is related to the multitask learning
framework~\cite{caruana1998multitask}. Despite its promise, this
framework has seen limited practical success in real world applications.
In speech recognition, there have been many successful reports of modeling
multiple languages using a single model (see~\cite{opac-b1119902} for an
extensive reference and references therein). Multilingual language
processing has also shown to be successful in domains other than
translation~\cite{GillickBVS15,TsvetkovSFLLMBL16}.

There have been other approaches similar to ours in spirit, but used for very
different purposes. In \cite{sennrich16polite}, the NMT framework has been
extended to control the politeness level of the target translation by adding a
special token to the source sentence. The same idea was used in
\cite{yamagishi16-WAT} to add the distinction between 'active' and 'passive'
tense to the generated target sentence.

Our method has an additional benefit not seen in other systems:
It gives the system the ability to perform zero-shot translation, meaning the
system can translate from a source language to a target
language without having seen explicit examples
from this specific language pair during training. Zero-shot translation
was the direct goal of~\cite{firat2016zero}. Although they were
not able to achieve this direct goal, they were able to do what they call
``zero-resource'' translation by using their pre-trained multi-way multilingual model
and later fine-tuning it with pseudo-parallel data generated by the model.
It should be noted that the difference between ``zero-shot'' and ``zero-resource''
translation is the additional fine-tuning step which is required in the latter
approach.

To the best of our knowledge, our work is the first to validate the use of
true multilingual translation using a single encoder-decoder model, and is
incidentally also already used in a production setting. It is also the first
work to demonstrate the possibility of zero-shot translation, a successful
example of transfer learning in machine translation, without any additional
steps.

\section{System Architecture for Multilingual Translation}
\label{system architecture}
\begin{figure}[h!]
\begin{center}
\centerline{\includegraphics[width=0.85\textwidth]{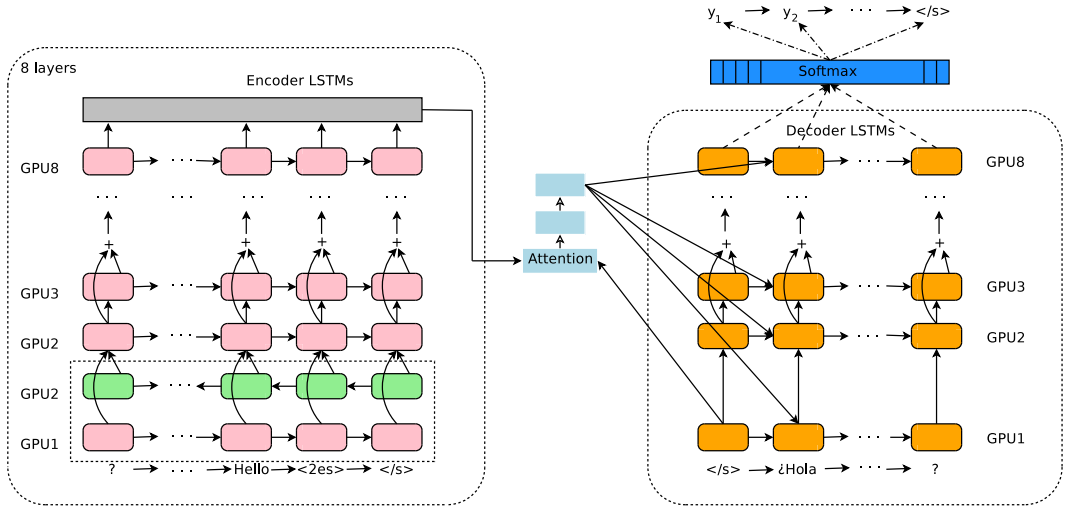}}
\caption{The model architecture of the Multilingual GNMT
  system.  In addition to what is described in~\cite{wu2016}, our
  input has an artificial token to indicate the required target language.
  In this example, the token ``<2es>''
  indicates that the target sentence is in Spanish, and the source
  sentence is reversed as a processing step. For most of our
  experiments we also used direct connections between the encoder and
  decoder although we later found out that the effect of these
  connections is negligible (however, once you train with those they
  have to be present for inference as well). The rest of the model
  architecture is the same as in~\cite{wu2016}.
}
\label{main_figure}
\end{center}
\end{figure}

The multilingual model architecture (see Figure~\ref{main_figure}) is
identical to Google's Neural Machine Translation (GNMT) system~\cite{wu2016}
(with the optional addition of direct connections between encoder and decoder
layers which we have used for some of our experiments, see description of
Figure~\ref{main_figure}) and we refer to that paper for a detailed description.

To be able to make use of multilingual data within a single system,
we propose one simple modification
to the input data, which is to introduce an artificial token at the
beginning of the input sentence to indicate the target language
the model should translate to. For instance, consider the following
English$\rightarrow$Spanish pair of sentences:
\begin{verbatim}
 Hello, how are you? -> Hola, ¿cómo estás?
\end{verbatim}
It will be modified to:
\begin{verbatim}
 <2es> Hello, how are you? -> Hola, ¿cómo estás?
\end{verbatim}
to indicate that Spanish is the target language. Note that we don't specify
the source language -- the model will learn this automatically. Not
specifying the source language has the potential disadvantage that words
with the same spelling but different meaning from different source
languages can be ambiguous to translate, but the advantage is that
it is simpler and we can handle input with code-switching. We find that
in almost all cases context provides enough language evidence to produce the
correct translation.

After adding the token to the input data, we train the model with
all multilingual data consisting of multiple language pairs at once, possibly
after over- or undersampling some of the data to adjust for the relative ratio
of the language data available. To address the issue of translation of unknown
words and to limit the vocabulary for computational efficiency, we use 
a shared wordpiece model ~\cite{wordpiece_schuster} across all the source and
target data used for training, usually with 32,000 word pieces. The
segmentation algorithm used here is very similar (with small differences)
to Byte-Pair-Encoding (BPE) which was described in
\cite{Gage:1994:NAD:177910.177914} and was also used in
\cite{SennrichHB15} for machine translation. Our training system is
implemented in Tensorflow~\cite{tensorflow2016}.
All training is carried out similar to~\cite{wu2016} and
implemented in TensorFlow~\cite{tensorflow2016}.

In summary, this approach is the simplest among the
alternatives that we are aware of. During training and inference, we
only need to add one additional token to each sentence of the source data to
specify the desired target language.

\section{Experiments and Results}
\label{experiments and results}
In this section, we apply our proposed method to train multilingual models
in several different configurations. Since we can have models
with either single or multiple source/target languages
we test three interesting cases for mapping languages:
\begin{itemize}
\item many source languages to one target language (many to one),
\item one source language to many target languages (one to many), and
\item many source languages to many target languages (many to many).
\end{itemize}
As already discussed in
Section~\ref{relwork}, other models have been used to explore
some of these cases already, but for completeness we apply our technique to
these interesting use cases again to give a full picture of the effectiveness
of our approach.

We will also show results and discuss benefits of bringing together
many (un)related languages in a single large-scale model trained on production
data. Finally, we will present our findings on zero-shot translation where the
model learns to translate between pairs of languages for which no explicit
parallel examples existed in the training data, and show results of experiments
where adding additional data improves zero-shot translation quality further.

\subsection{Datasets, Training Protocols and Evaluation Metrics}
For WMT, we train our models on the WMT'14 English(En)$\rightarrow$French(Fr)
and the WMT'14 English$\rightarrow$German(De) datasets.
In both cases, we use newstest2014 as the test sets
to be able to compare against previous
work~\cite{luong2015addressing,jean2015using,DBLP:journals/corr/ZhouCWLX16,wu2016}.
For WMT Fr$\rightarrow$En and De$\rightarrow$En we use newstest2014
and newstest2015 as test sets. Despite training on WMT'14 data, which is
somewhat smaller than WMT'15, we test our De$\rightarrow$En model
on newstest2015, similar to~\cite{luong-pham-manning:2015:EMNLP}.
The combination of newstest2012 and newstest2013 is used as the
development set.

In addition to WMT, we also evaluate the multilingual approach on some
Google-internal large-scale production datasets representing a wide
spectrum of languages with very distinct
linguistic properties: English$\leftrightarrow$Japanese(Ja),
English$\leftrightarrow$Korean(Ko), English$\leftrightarrow$Spanish(Es), and
English$\leftrightarrow$Portuguese(Pt). These datasets are two to three
orders of magnitude larger than the WMT datasets.

Our training protocols are mostly identical to those described
in~\cite{wu2016} and we refer the reader to the detailed description
in that paper.
We find that some multilingual models take a little more time to train than
single language pair models,
likely because each language pair is seen only for a fraction of the training
process. Depending on the number of languages a full
training can take up to 10M steps and 3 weeks to converge (on roughly 100
GPUs). We use larger batch sizes with a slightly higher initial learning rate
to speed up the convergence of these models.

We evaluate our models using the standard BLEU score metric and to
make our results comparable
to~\cite{sutskever2014sequence,luong2015addressing,DBLP:journals/corr/ZhouCWLX16,wu2016},
we report tokenized BLEU score as computed by the
\texttt{multi-bleu.pl} script, which can be downloaded from the public
implementation of Moses.\footnote{\url{http://www.statmt.org/moses/}}

To test the influence of varying amounts of training data per language pair
we explore two strategies when building multilingual models: a)
where we oversample the data from all language pairs to be of the same size
as the largest language pair, and b) where we mix the data as is without
any change. The wordpiece model training is done after the optional oversampling
taking into account all the changed data ratios.
For the WMT models we report results using both of these strategies.
For the production models, we always balance the data such that the ratios
are equal.

One benefit of the way we share all the components of the model is
that the mini-batches can contain data from different language pairs during
training and inference, which are typically just random samples from the
final training and test data distributions. This is a simple way of preventing
``catastrophic forgetting'' - tendency for knowledge of previously learnt
task(s) (e.g. language pair A) to be abruptly forgotten as information relevant
to the current task (e.g. language pair B) is incorporated \cite{french1999catastrophic}.
Other approaches to multilingual translation require complex
update scheduling mechanisms to prevent this effect \cite{firatjournal}.

\subsection{Many to One}
In this section we explore having multiple source languages and a
single target language --- the simplest way of combining language pairs.
Since there is only a single target language no additional source token is
required. We perform three sets of experiments:
\begin{itemize}
\item The first set of experiments is on the WMT datasets, where we
  combine German$\rightarrow$English and French$\rightarrow$English
  to train a multilingual model. Our baselines are two single language pair
  models: German$\rightarrow$English and French$\rightarrow$English
  trained independently. We perform these experiments once with
  oversampling and once without.
\item The second set of experiments is
  on production data where we combine Japanese$\rightarrow$English
  and Korean$\rightarrow$English, with oversampling. The baselines are two
  single language pair models: Japanese$\rightarrow$English and
  Korean$\rightarrow$English trained independently.
\item Finally, the third set of experiments is on
  production data where we combine Spanish$\rightarrow$English and
  Portuguese$\rightarrow$English, with oversampling. The baselines are again
  two single language pair models trained independently.
\end{itemize}
All of the multilingual and single language pair models have the same total
number of parameters as the baseline NMT models trained on a single language
pair (using 1024 nodes, 8 LSTM layers and
a shared wordpiece model vocabulary of 32k, a total of 255M parameters per
model). A side effect of this equal choice of parameters is that it is
presumably unfair to the multilingual models as the number of parameters
available per language pair is reduced by a factor of $N$ compared to the single
language pair models, if $N$ is the number of language pairs combined in the
multilingual model. The multilingual model also has to handle the combined
vocabulary of all the single models. We chose to keep the number of
parameters constant for all models to simplify experimentation. We relax this
constraint for some of the large-scale experiments shown further below.

\begin{table}[h!]
\caption{Many to One: BLEU scores on various data sets for single language pair and multilingual models.}
\label{table:many_to_one}
\centering
\begin{tabular}{r c c c }
\hline\hline 
Model & Single   & Multi  & Diff \\\hline
WMT German$\rightarrow$English  (oversampling) & 30.43 & 30.59 & +0.16 \\ 
WMT French$\rightarrow$English  (oversampling) & 35.50 & 35.73 & +0.23 \\
\hline
WMT German$\rightarrow$English  (no oversampling) & 30.43 & 30.54 & +0.11 \\ 
WMT French$\rightarrow$English  (no oversampling) & 35.50 & 36.77 & +1.27 \\
\hline\hline
Prod Japanese$\rightarrow$English  & 23.41 & 23.87 & +0.46 \\
Prod Korean$\rightarrow$English  & 25.42 & 25.47 & +0.05 \\
\hline\hline
Prod Spanish$\rightarrow$English  & 38.00 & 38.73 & +0.73 \\
Prod Portuguese$\rightarrow$English  & 44.40 & 45.19 & +0.79 \\
\hline\hline
\end{tabular}
\end{table}

The results are presented in Table~\ref{table:many_to_one}. For all
experiments the multilingual models outperform the baseline single systems
despite the above mentioned disadvantage with respect to the
number of parameters available per language pair. One possible
hypothesis explaining the gains is that the model has been shown more English
data
on the target side, and that the source languages belong to the same
language families, so the model has learned useful generalizations.

For the WMT experiments, we obtain a maximum gain of +1.27
BLEU for French$\rightarrow$English. Note that the results on both
the WMT test sets are better than other published state-of-the-art results
for a single model, to the best of our knowledge.
On the production experiments, we see that the multilingual models
outperform the baseline single systems by as much as +0.8 BLEU.

\subsection{One to Many}
In this section, we explore the application of our method when there is a
single source language and multiple target languages. Here we need to prepend
the input with an additional token to specify the target language. We perform
three sets of experiments almost identical to the previous section except
that the source and target languages have been reversed.

Table~\ref{table:one_to_many} summarizes the results when performing
translations into multiple target languages. We see that the
multilingual models are comparable to, and in some cases
outperform, the baselines, but not always. We obtain a large gain of
+0.9 BLEU for English$\rightarrow$Spanish. Unlike the previous set of
results, there are less significant gains in this set of experiments.
This is perhaps due to the fact that the decoder has a more difficult time
translating into multiple target languages which may even have different
scripts, which are combined into a single shared
wordpiece vocabulary. Note that even for languages with entirely different
scripts (e.g. Korean and Japanese) there is significant overlap in wordpieces
when real data is used, as often numbers, dates, names, websites, punctuation
etc. are actually using a shared script (ASCII).

\begin{table}[h!]
\caption{One to Many: BLEU scores on various data sets for single language pair and multilingual models.}
\label{table:one_to_many}
\centering
\begin{tabular}{r c c c }
\hline\hline 
Model & Single   & Multi  & Diff \\\hline
WMT English$\rightarrow$German (oversampling)     & 24.67 & 24.97 & +0.30 \\
WMT English$\rightarrow$French (oversampling)     & 38.95 & 36.84 & -2.11 \\
\hline
WMT English$\rightarrow$German (no oversampling)  & 24.67 & 22.61 & -2.06 \\
WMT English$\rightarrow$French (no oversampling)  & 38.95 & 38.16 & -0.79 \\
\hline\hline
Prod English$\rightarrow$Japanese   & 23.66 & 23.73 & +0.07 \\
Prod English$\rightarrow$Korean     & 19.75 & 19.58 & -0.17 \\
\hline\hline
Prod English$\rightarrow$Spanish    & 34.50 & 35.40 & +0.90 \\
Prod English$\rightarrow$Portuguese & 38.40 & 38.63 & +0.23 \\
\hline\hline
\end{tabular}
\end{table}

We observe that oversampling helps the smaller language pair (En$\rightarrow$De)
at the cost of lower quality for the larger language pair (En$\rightarrow$Fr).
The model without oversampling achieves better results on the larger language
compared to the smaller one as expected. We also find that this effect
is more prominent on smaller datasets (WMT) and much less so on our much larger
production datasets.


\subsection{Many to Many}
In this section, we report on experiments
when there are multiple source languages and multiple target
languages within a single model --- the most difficult setup. Since multiple
target languages are given, the input needs to be prepended with the target
language token as above.

The results are presented in Table~\ref{table:many_to_many}. We see that
the multilingual production models with the same model size and vocabulary size
as the single language models are quite close to the baselines --
the average relative loss in BLEU score across all experiments is only
approximately 2.5\%.

\begin{table}[h!]
\caption{Many to Many: BLEU scores on various data sets for single language pair and multilingual models.}
\label{table:many_to_many}
\centering
\begin{tabular}{r c c c }
\hline\hline 
Model & Single   & Multi  & Diff  \\\hline
WMT English$\rightarrow$German (oversampling)    & 24.67  & 24.49 & -0.18 \\
WMT English$\rightarrow$French (oversampling)    & 38.95 & 36.23 & -2.72 \\
WMT German$\rightarrow$English (oversampling)    & 30.43 & 29.84 & -0.59 \\
WMT French$\rightarrow$English (oversampling)    & 35.50 & 34.89 & -0.61 \\
\hline
WMT English$\rightarrow$German (no oversampling) & 24.67 & 21.92 & -2.75 \\
WMT English$\rightarrow$French (no oversampling) & 38.95 & 37.45 & -1.50 \\
WMT German$\rightarrow$English (no oversampling) & 30.43 & 29.22 & -1.21 \\
WMT French$\rightarrow$English (no oversampling) & 35.50 & 35.93 & +0.43 \\
\hline\hline
Prod English$\rightarrow$Japanese    & 23.66 & 23.12 & -0.54 \\
Prod English$\rightarrow$Korean      & 19.75 & 19.73 & -0.02 \\
Prod Japanese$\rightarrow$English    & 23.41 & 22.86 & -0.55 \\
Prod Korean$\rightarrow$English      & 25.42 & 24.76 & -0.66 \\
\hline\hline
Prod English$\rightarrow$Spanish     & 34.50 & 34.69 & +0.19 \\
Prod English$\rightarrow$Portuguese  & 38.40 & 37.25 & -1.15 \\
Prod Spanish$\rightarrow$English     & 38.00 & 37.65 & -0.35 \\
Prod Portuguese$\rightarrow$English  & 44.40 & 44.02 & -0.38 \\
\hline\hline
\end{tabular}
\end{table}

On the WMT datasets, we once again explore the impact of oversampling
the smaller language pairs. We notice a similar trend to the previous
section in which oversampling helps the smaller language pairs at the
expense of the larger ones, while not oversampling seems to have the
reverse effect.

Although there are some significant losses in quality
from training many languages jointly using a model with the same total
number of parameters as the single language pair models, these models
reduce the total complexity involved in training and productionization.
Additionally, these multilingual models have more interesting advantages as will be
discussed in more detail in the sections below.

\subsection{Large-scale Experiments}
This section shows the result of combining 12 production language pairs having
a total of 3B parameters (255M per single model) into a single multilingual
model. A range of multilingual models were trained, starting from the same size
as a single language pair model with 255M parameters (1024 nodes) up to 650M
parameters (1792 nodes).
As above, the input needs to be prepended with the target language token.
We oversample the examples from the smaller language pairs to balance the
data as explained above.

The results for single language pair models versus multilingual models with
increasing numbers of parameters are summarized in
Table~\ref{table:large_scale}. We find that the multilingual models are on
average worse than the single models (about 5.6\% to 2.5\% relative depending
on size, however, some actually get better) and as expected the average
difference gets smaller when going to larger
multilingual models. It should be noted that the largest multilingual model we have trained has
still about five times less parameters than the combined single models.

The multilingual model also requires only roughly 1/12-th of the training
time (or computing resources) to converge compared to the combined single
models (total training time for all our models is still in the
order of weeks). Another important point is that since we only train for a
little longer than a standard single model, the individual language pairs can
see as little as 1/12-th of the data in comparison to their single language
pair models but still produce satisfactory results.

\begin{table}[h!]
\caption{Large-scale experiments: BLEU scores for single language pair and multilingual models.}
\label{table:large_scale}
\centering
\begin{tabular}{c|c|c c c c}
  \hline\hline 
  Model             & Single & Multi & Multi & Multi & Multi  \\
  \hline
  \#nodes           & 1024   & 1024  & 1280  & 1536  & 1792 \\
  \#params          & 3B     & 255M  & 367M  & 499M  & 650M  \\
  \hline
  Prod English$\rightarrow$Japanese & 23.66  & 21.10 & 21.17 & 21.72 & 21.70 \\
  Prod English$\rightarrow$Korean & 19.75  & 18.41 & 18.36 & 18.30 & 18.28 \\
  Prod Japanese$\rightarrow$English & 23.41  & 21.62 & 22.03 & 22.51 & 23.18 \\
  Prod Korean$\rightarrow$English & 25.42  & 22.87 & 23.46 & 24.00 & 24.67 \\
  Prod English$\rightarrow$Spanish & 34.50  & 34.25 & 34.40 & 34.77 & 34.70 \\
  Prod English$\rightarrow$Portuguese & 38.40  & 37.35 & 37.42 & 37.80 & 37.92 \\
  Prod Spanish$\rightarrow$English & 38.00  & 36.04 & 36.50 & 37.26 & 37.45 \\
  Prod Portuguese$\rightarrow$English & 44.40  & 42.53 & 42.82 & 43.64 & 43.87 \\
  Prod English$\rightarrow$German & 26.43  & 23.15 & 23.77 & 23.63 & 24.01 \\
  Prod English$\rightarrow$French & 35.37  & 34.00 & 34.19 & 34.91 & 34.81 \\
  Prod German$\rightarrow$English & 31.77  & 31.17 & 31.65 & 32.24 & 32.32 \\
  Prod French$\rightarrow$English & 36.47  & 34.40 & 34.56 & 35.35 & 35.52 \\
  \hline
  ave diff          &  -     & -1.72 & -1.43 & -0.95 & -0.76 \\
  vs single         &  -     & -5.6\% & -4.7\% & -3.1\% & -2.5\% \\
  \hline\hline
\end{tabular}
\end{table}
The results are summarized in Table~\ref{table:large_scale}. We find that the
multilingual model is reasonably close to the best single models and in some
cases even achieves comparable quality. It is remarkable that a single model
with 255$M$ parameters can do what 12 models with a total of 3$B$ parameters
would have done. The multilingual model also requires one twelfth of the training
time and computing resources to converge. Another important point
is that since we only train for a little longer than the single models,
the individual language pairs can see as low as one twelfth of the data in comparison
to their single language pair models. Again we note that the comparison below is
somewhat unfair for the multilingual
model and we expect a larger model trained on all available data will
likely achieve comparable or better quality than the baselines.

In summary, multilingual NMT enables us to group languages with little or no
loss in quality while having the benefits of better training efficiency, smaller
number of models, and easier productionization.

\subsection{Zero-Shot Translation}
The most straight-forward approach of translating between languages where no
or little parallel data is available is to use explicit bridging,
meaning to translate to an intermediate language first and
then to translate to the desired target language. The intermediate language is
often English as xx$\rightarrow$en and en$\rightarrow$yy data is more readily
available. The two potential
disadvantages of this approach are: a) total translation time doubles, b) the
potential loss of quality by translating to/from the intermediate language.

An interesting benefit of our approach is that it allows to perform directly
{\em implicit bridging} (zero-shot translation) between a language pair for
which no explicit parallel training data has been seen without any
modification to the model.
Obviously, the model will only be able to do zero-shot translation
between languages it has seen individually as source and target
languages during training at some point, not for entirely new ones.

To demonstrate this we will use two multilingual models --- a model trained
with examples from two different language pairs,
Portuguese$\rightarrow$English and English$\rightarrow$Spanish (Model 1), and
a model trained with examples from four different language pairs,
English$\leftrightarrow$Portuguese and English$\leftrightarrow$Spanish
(Model 2). We show that both of these models can generate
reasonably good quality Portuguese$\rightarrow$Spanish
translations (BLEU scores above 20) without
ever having seen Portuguese$\rightarrow$Spanish data during training. To our
knowledge this is the first demonstration of true multilingual zero-shot
translation. As with the previous multilingual models, both of these
models perform comparable to or even slightly better than the
baseline single language pair models. Note that besides the pleasant fact that
zero-shot translation works at all it has also the advantage of halving
decoding speed as no explicit bridging through a third language
is necessary when translating from Portuguese to Spanish.

Table~\ref{table:zrt} summarizes our results for the
Portuguese$\rightarrow$Spanish translation experiments. Rows (a) and (b)
report the performance of the phrase-based machine translation (PBMT)
system and the NMT system
through bridging (translation from Portuguese to English and
translating the resulting English sentence to Spanish). It can be seen that
the NMT system outperforms the PBMT system by close to 2 BLEU points. Note
that Model 1 and Model 2 can be bridged within themselves to perform
Portuguese$\rightarrow$Spanish translation. We do not report these numbers
since they are similar to the performance of bridging with two individual
single language pair NMT models.
For comparison, we built a single NMT model on all available
Portuguese$\rightarrow$Spanish parallel sentences (see (c) in
Table~\ref{table:zrt}).


\begin{table}[h!]
  \caption{Portuguese$\rightarrow$Spanish BLEU scores using various models.}
  \label{table:zrt}
  \centering
  \begin{tabular}{r c c c c }
    \hline\hline 
    & Model & Zero-shot & BLEU \\\hline
    (a) & PBMT bridged & no &  28.99 \\
    (b) & NMT bridged & no & 30.91 \\
    (c) & NMT Pt$\rightarrow$Es & no & 31.50 \\
    (d) & Model 1 (Pt$\rightarrow$En, En$\rightarrow$Es) & yes & 21.62 \\
    (e) & Model 2 (En$\leftrightarrow$\{Es, Pt\}) & yes & 24.75 \\
    (f) & Model 2 + incremental training & no & 31.77 \\
    \hline 
  \end{tabular}
\end{table}

The most interesting observation is that both Model 1 and Model 2 can perform
zero-shot translation with reasonable quality (see (d) and (e)) compared to
the initial expectation that this would not work at all. Note that Model 2
outperforms Model 1 by close to 3 BLEU points although
Model 2 was trained with four language pairs as opposed to with only two
for Model 1 (with both models having the same number of total parameters). In
this case the addition of Spanish on the source side and
Portuguese on the target side helps Pt$\rightarrow$Es zero-shot translation
(which is the opposite direction of where we would expect it to help).
We believe that this unexpected effect is only possible because our shared
architecture enables the model to learn a form of interlingua between all
these languages. We explore this hypothesis in more detail in
Section~\ref{visualanalysis}.

Finally we incrementally train zero-shot Model 2 with a small amount of true
Pt$\rightarrow$Es parallel data (an order of magnitude less than
Table~\ref{table:zrt} (c)) and obtain the best quality and half the decoding
time compared to explicit bridging (Table~\ref{table:zrt} (b)). The resulting
model cannot be called zero-shot anymore since some true parallel data has been
used to improve it. Overall this shows that the proposed approach of implicit
bridging using zero-shot
translation via multilingual models can serve as a good baseline for further
incremental training with relatively small amounts of true parallel data of
the zero-shot direction. This result is especially significant for non-English
low-resource language pairs where it might be easier to obtain parallel data
with English but much harder to obtain parallel data for language pairs where
neither the source nor the target language is English. We explore the effect
of using parallel data in more detail in Section~\ref{parallel-data}.

Since Portuguese and Spanish are of the same language family an interesting
question is how well zero-shot translation works for less related languages.
Table~\ref{table:zrt2} shows the results for explicit and implicit bridging from
Spanish to Japanese using the large-scale model from
Table~\ref{table:large_scale} -- Spanish and Japanese can be regarded as
quite unrelated. As expected zero-shot translation works worse
than explicit bridging and the quality drops relatively more (roughly 50\%
drop in BLEU score) than for the case of more related languages as shown above.
Despite the quality drop, this proves that our approach enables zero-shot
translation even between unrelated languages.

\begin{table}[h!]
  \caption{Spanish$\rightarrow$Japanese BLEU scores for explicit and implicit
  bridging using the 12-language pair large-scale model from
  Table~\ref{table:large_scale}.}
  \label{table:zrt2}
  \centering
  \begin{tabular}{c c }
    \hline\hline 
    Model & BLEU \\\hline
    NMT Es$\rightarrow$Ja explicitly bridged & 18.00 \\
    NMT Es$\rightarrow$Ja implicitly bridged & 9.14 \\
    \hline 
  \end{tabular}
\end{table}

\subsection{Effect of Direct Parallel Data}
\label{parallel-data}
In this section, we explore two ways of leveraging available parallel
data to improve zero-shot translation quality, similar in spirit to what was
reported in \cite{firat2016zero}. For
our multilingual architecture we consider:
\begin{itemize}
\item Incrementally training the multilingual model on the additional parallel
  data for the zero-shot directions.
\item Training a new multilingual model with all available parallel data mixed
  equally.
\end{itemize}
For our experiments, we use a baseline model which we call
``Zero-Shot'' trained on a combined parallel corpus of
English$\leftrightarrow$\{Belarusian(Be), Russian(Ru), Ukrainian(Uk)\}.
We trained a second model on the above corpus together with additional
Ru$\leftrightarrow$\{Be, Uk\} data.
We call this model ``From-Scratch''.
Both models support four target languages, and are evaluated on our standard
test sets. As done previously we oversample the data such that all language
pairs are represented equally. Finally, we take the best checkpoint of the
``Zero-Shot'' model, and run incremental training on a small portion
of the data used to train the ``From-Scratch'' model for a short period of
time until convergence (in this case 3\% of ``Zero-Shot'' model total training
time). We call this model ``Incremental''.

As can be seen from Table~\ref{table:enrubeuk_scores},
for the English$\leftrightarrow$X directions,
all three models show comparable scores.
On the Russian$\leftrightarrow$\{Belarusian, Ukrainian\} directions,
the ``Zero-Shot'' model already achieves relatively high BLEU scores
for all directions except one, without any explicit parallel data.
This could be because these languages are linguistically related.
In the ``From-Scratch'' column, we see that training
a new model from scratch improves the zero-shot translation directions
further. However, this strategy has a slightly negative effect on the
English$\leftrightarrow$X directions because our oversampling strategy
will reduce the frequency of the data from these directions. In the final
column, we see that incremental training with direct parallel data
recovers most of the BLEU score difference between the first two columns
on the zero-shot language pairs. In summary, our shared architecture
models the zero-shot language pairs quite well and hence enables us to
easily improve their quality with a small amount of additional parallel data.

\begin{table}[h!]
\caption{BLEU scores for English$\leftrightarrow$\{Belarusian, Russian, Ukrainian\} models.}
\label{table:enrubeuk_scores}
\centering
\begin{tabular}{r c c c }
\hline\hline
& Zero-Shot &  From-Scratch & Incremental \\
\hline
English$\rightarrow$Belarusian & 16.85 & 17.03 & 16.99  \\
English$\rightarrow$Russian    & 22.21 & 22.03 & 21.92  \\
English$\rightarrow$Ukrainian  & 18.16 & 17.75 & 18.27  \\
Belarusian$\rightarrow$English & 25.44 & 24.72 & 25.54  \\
Russian$\rightarrow$English    & 28.36 & 27.90 & 28.46  \\
Ukrainian$\rightarrow$English  & 28.60 & 28.51 & 28.58  \\
\hline
Belarusian$\rightarrow$Russian & 56.53 & 82.50 & 78.63  \\
Russian$\rightarrow$Belarusian & 58.75 & 72.06 & 70.01  \\
Russian$\rightarrow$Ukrainian  & 21.92 & 25.75 & 25.34  \\
Ukrainian$\rightarrow$Russian  & 16.73 & 30.53 & 29.92  \\
\hline
\end{tabular}
\end{table}

\section{Visual Analysis}
\label{visualanalysis}
The results of this paper --- that training a model across multiple languages
can enhance performance at the individual language level, and that zero-shot
translation can be effective --- raise a number of questions about how
these tasks are handled inside the model, for example:
\begin{itemize}
\item Is the network learning some sort of shared representation, in which
sentences with the same meaning are represented in similar ways regardless of
language?
\item Does the model operate on zero-shot translations in the
same way as it treats language pairs it has been trained on?
\end{itemize}

One way to study the representations used by the network is to look at the
activations of the network during translation. A starting point for
investigation is the set of {\it context vectors}, i.e.,
the sum of internal encoder states weighted by their attention probabilities
per step (Eq. (5) in \cite{BahdanauCB15}).

A translation of a single sentence generates a sequence of context vectors. In
this context, our original questions about shared representation can be
studied by looking at how the vector sequences of different sentences
relate. We could then ask for example:
\begin{itemize}
\item Do sentences cluster together depending on the source or target language?
\item Or instead do sentences with similar meanings cluster, regardless of language?
\end{itemize}
We try to find answers to these questions by looking at lower-dimensional
representations of internal embeddings of the network that humans can more
easily interpret.

\subsection{Evidence for an Interlingua}
\begin{figure}[h!]
  \includegraphics[width=1.0\textwidth]{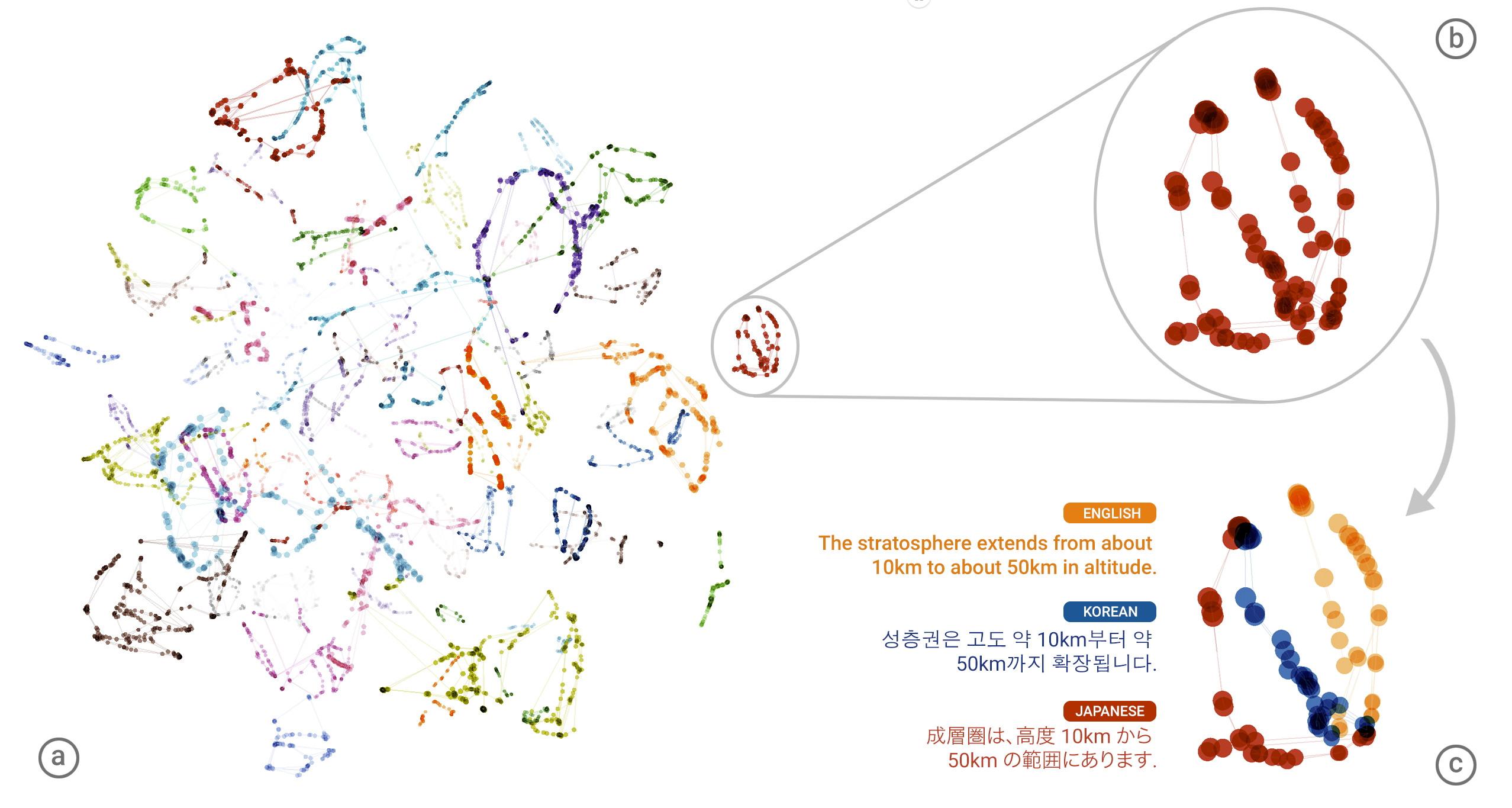}
  \caption{A t-SNE projection of the embedding of 74 semantically identical
    sentences translated across all 6 possible directions, yielding a total of
    9,978 steps (dots in the image),
    from the model trained on English$\leftrightarrow$Japanese and
    English$\leftrightarrow$Korean examples. (a) A bird's-eye view of the
    embedding, coloring by the index of the semantic sentence. Well-defined
    clusters each having a single color are apparent. (b) A zoomed in view
    of one of the clusters with the same coloring. All of the sentences within
    this cluster are translations of ``The stratosphere extends from about 10km
    to about 50km in altitude.'' (c) The same cluster colored by source
    language. All three source languages can be seen within this cluster.}
\label{enjako_embedding}
\end{figure}

Several trained networks indeed show strong visual evidence of a shared
representation. For example, Figure~\ref{enjako_embedding} below was
produced from
a many-to-many model trained on four language pairs,
English$\leftrightarrow$Japanese and
English$\leftrightarrow$Korean. To visualize the model in action we began
with a small corpus of 74 triples of semantically identical cross-language
phrases. That is, each triple contained phrases in English,
Japanese and Korean with the same underlying meaning. To compile
these triples, we searched a ground-truth database for
English sentences which were paired with both Japanese and Korean translations.

We then applied the trained model to translate each sentence of each triple into
the two other possible languages. Performing this process yielded six new
sentences based on each triple, for a total of $74  * 6 = 444$ total
translations with 9,978 steps corresponding to the same number of context
vectors. Since context vectors are high-dimensional, we use the TensorFlow
Embedding Projector%
\footnote{\url{https://www.tensorflow.org/get_started/embedding_viz}}
to map them into more accessible 3D space via t-SNE
\cite{Maaten2008}. In the following diagrams, each point represents a single
decoding step during the translation process. Points that represent steps for
a given sentence are connected by line segments.

Figure~\ref{enjako_embedding} shows a global view of all 9,978 context
vectors. Points produced
from the same original sentence triple are all given the same (random) color.
Inspection of these clusters shows that each strand
represents a single sentence, and clusters of strands generally represent
a set of translations of the same underlying sentence, but with different
source and target languages.

At right are two close-ups: one of an individual cluster, still coloring based
on membership in the same triple, and one where we have colored by source
language.

\subsection{Partially Separated Representations}
Not all models show such clean semantic clustering. Sometimes we observed
joint embeddings in some regions of space coexisting with
separate large clusters which contained many context vectors from
just one language pair.

\begin{figure}[h!]
  \includegraphics[width=1\textwidth]{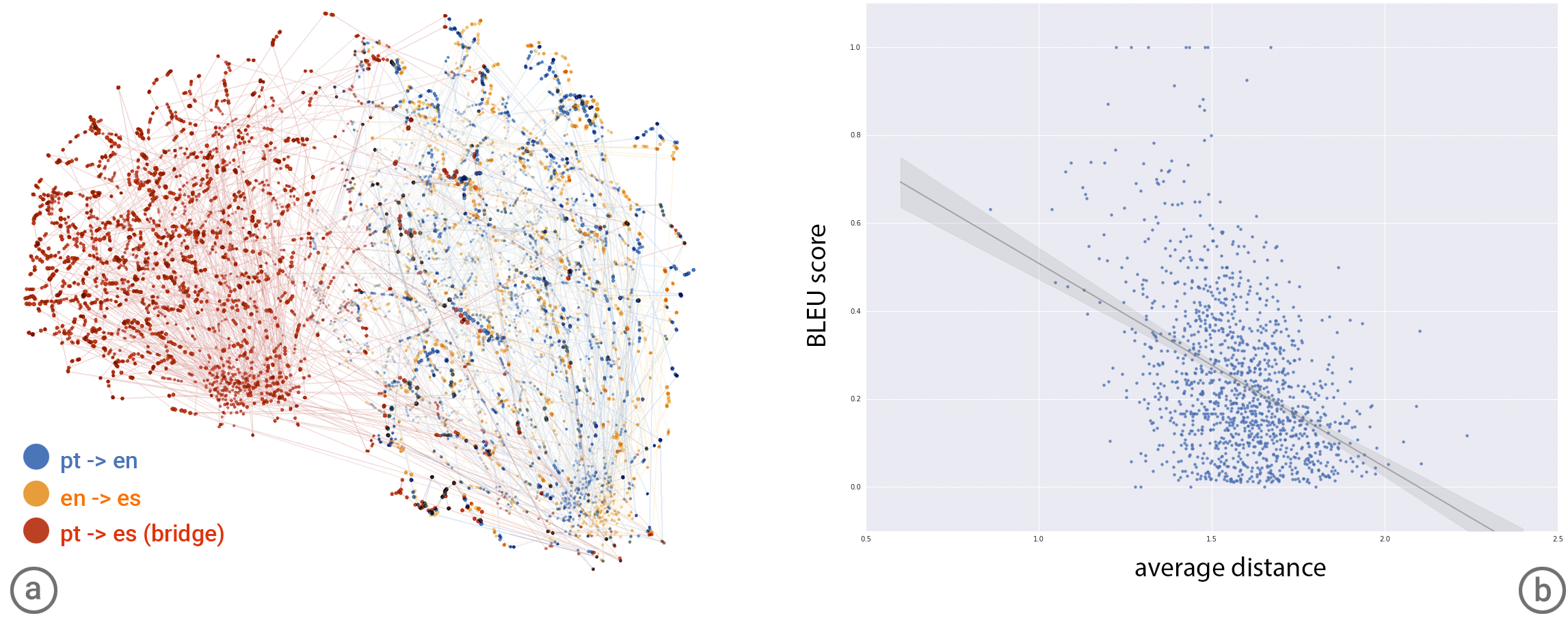}
  \caption{(a) A bird's-eye view of a t-SNE projection of an embedding of the
    model trained on Portuguese$\rightarrow$English (blue) and
    English$\rightarrow$Spanish (yellow) examples with a
    Portuguese$\rightarrow$Spanish zero-shot bridge (red). The large red
    region on the left primarily contains the zero-shot
    Portuguese$\rightarrow$Spanish translations. (b) A scatter plot of BLEU
    scores of zero-shot translations versus the average point-wise distance
    between the zero-shot translation and a non-bridged translation. The Pearson
    correlation coefficient is $-0.42$.}
\label{enespt_embedding_and_bleu_scatter}
\end{figure}

For example, Figure~\ref{enespt_embedding_and_bleu_scatter}a shows a t-SNE
projection of context vectors from a model that was trained on
Portuguese$\rightarrow$English (blue) and English$\rightarrow$Spanish (yellow)
and performing zero-shot translation from Portuguese$\rightarrow$Spanish
(red). This projection shows 153 semantically identical triples translated
as described above, yielding 459 total translations.
The large red region on the left primarily contains zero-shot
Portuguese$\rightarrow$Spanish translations. In other words, for a significant
number of sentences, the zero-shot translation has a different embedding than
the two trained translation directions. On the other hand, some zero-shot
translation vectors do seem to fall near the embeddings found in other
languages, as on the large region on the right.

It is natural to ask whether the large cluster of ``separated'' zero-shot
translations has any significance. A definitive answer requires further
investigation, but in this case zero-shot translations in the separated area do
tend to have lower BLEU scores.

To measure the relationship between translation quality and distance between
embeddings of the same semantic sentence, we first calculated BLEU scores
for each translation. (This is possible since all triples of phrases were
extracted from ground truth data.) Next, we needed to define a dissimilarity
measure for embeddings of different sentences, accounting for the fact
that two sentences
might consist of different numbers of wordpieces. To do so, for a sentence
of $n$ wordpieces $w_0, w_1, \dots, w_{n-1}$ where
the $i$\,th wordpiece has been embedded at $y_i \in \mathbb{R}^{1024}$, we
defined a curve $\gamma: [0,1] \to \mathbb{R}^{1024}$ at ``control points''
of the form $\frac{i}{n-1}$ by:
$$\gamma\Big(\frac{i}{n-1}\Big) = y_i$$
and use linear interpolation to define $\gamma$ between these points. The
dissimilarity between two curves $\gamma_1$ and $\gamma_2$, where
$m$ is the maximum number of wordpieces in both sentences, is defined by
$$dissimilarity(\gamma_1, \gamma_2) =
\frac{1}{m} \sum_{i=0}^{m-1} d\bigg(\gamma_1\Big(\frac{i}{m-1}\Big),
                                    \gamma_2\Big(\frac{i}{m-1}\Big)\bigg)$$

Figure~\ref{enespt_embedding_and_bleu_scatter}b shows a plot of BLEU scores of
a zero-shot translation versus the average pointwise distance between it and the
same translation from a trained language pair. We can see that the value
of this dissimilarity score is correlated with the quality of the
zero-shot translation with a Pearson correlation coefficient of $-0.42$,
indicating moderate correlation. An interesting
area for future research is to find a more reliable correspondence between
embedding geometry and model performance to predict the quality of a zero-shot
translation during decoding by comparing it to the embedding of the translation
through a trained language pair.

\section{Mixing Languages}
\label{mixinglanguages}
\newcommand{\ja}[1]{\begin{CJK}{UTF8}{min}#1\end{CJK}}
\newcommand{\ko}[1]{\begin{CJK}{UTF8}{mj}#1\end{CJK}}

Having a mechanism to translate from a random source language to a single
chosen target language using an additional source token made us think about
what happens when languages are mixed on the source or target side.
In particular, we were interested in the following two experiments:
\begin{enumerate}
\item Can a multilingual model successfully handle multi-language input
(code-switching), when it happens in the middle of the sentence?
\item What happens when a multilingual model is triggered not with a single but
two target language tokens weighted such that their weight adds up to one (the
equivalent of merging the weighted embeddings of these tokens)?
\end{enumerate}
The following two sections discuss these experiments.

\subsection{Source Language Code-Switching}
In this section we show how multilingual models deal with source language
code-switching. Here we show an example from a multilingual model
that was trained with Japanese,Korean$\rightarrow$English data.
Using this model, mixing Japanese and Korean in the source
produces in many cases correct English translations, showing the code-switching
can be handled by this model, although no such code-switching samples were
present in the training data. Note that the model can effectively
handle the different typographic scripts since the individual characters/wordpieces
are present in our wordpiece vocabulary.
\begin{itemize}
\item {\bf Japanese: }\ja{私は東京大学の学生です。} $\rightarrow$ I am a student at Tokyo University.
\item {\bf Korean: }\ko{나는 도쿄 대학의 학생입니다. } $\rightarrow$ I am a student at Tokyo University.
\item {\bf Mixed Japanese/Korean: }\ja{私は東京大学}\ko{학생입니다.} $\rightarrow$ I am a student of Tokyo University.
\end{itemize}
Interestingly, the translation for the mixed-language input differs slightly
from both of the single source language translations. In practice, it is
not too hard to find examples where code-switching in the input does not
result in good outputs; in some cases the model will simply copy parts of the
source sentence instead of translating it.

\subsection{Weighted Target Language Selection}
In this section we test what happens when we mix target languages.
We take a multilingual model trained with multiple target languages,
for example, English$\rightarrow$\{Japanese, Korean\}.
Then instead of feeding the embedding vector for ``<2ja>'' to the
bottom layer of the encoder LSTM, we feed a linear combination
$(1-w)\mbox{<2ja>} + w\mbox{<2ko>}$.
Clearly, for $w=0$ the model should produce Japanese,
for $w=1$ it should produce Korean, but what happens in between?

One expectation could be that the model will output some sort of intermediate
language (``Japarean''), but the results turn out to be less surprising.
Most of the time the output just switches from one language to another around $w=0.5$.
In some cases, for intermediate values of $w$ the model switches languages mid-sentence.

A possible explanation for this behavior is that the target language model,
implicitly learned by the decoder LSTM, may make it very hard to mix words
from different languages, especially when these languages use different scripts.
In addition, since the token which defines the requested target language is
placed at the beginning of the sentence, the further the decoder progresses,
the less likely it is to put attention on this token, and instead the choice
of language is determined by previously generated target words.

\begin{table}[h!]
  \caption{Several examples of gradually mixing target languages in multilingual models.}
  \label{table:mixlangexamples}
\begin{tabular}{ c l }
\rule{0pt}{3ex}
Russian/Belarusian: & I wonder what they’ll do next!\\
\hline
$w_{be}=0.00$ & \foreignlanguage{russian}{Интересно, что они сделают дальше!} \\
$w_{be}=0.20$ & \foreignlanguage{russian}{Интересно, что они сделают дальше!} \\
$w_{be}=0.30$ & \foreignlanguage{russian}{\underline{Цікаво}, что они будут делать дальше!} \\
$w_{be}=0.44$ & \foreignlanguage{russian}{\underline{Цікаво, що вони будуть робити далі!}} \\
$w_{be}=0.46$ & \foreignlanguage{russian}{\underline{Цікаво, що вони будуть робити далі!}} \\
$w_{be}=0.48$ & \foreignlanguage{russian}{\underline{Цікаво}, што яны зробяць далей!} \\
$w_{be}=0.50$ & \foreignlanguage{russian}{Цікава, што яны будуць рабіць далей!} \\
$w_{be}=1.00$ & \foreignlanguage{russian}{Цікава, што яны будуць рабіць далей!} \\
\rule{0pt}{3ex}
Japanese/Korean: & I must be getting somewhere near the centre of the earth. \\
\hline
$w_{ko}=0.00$ & \ja{私は地球の中心の近くにどこかに行っているに違いない。} \\
$w_{ko}=0.40$ & \ja{私は地球の中心近くのどこかに着いているに違いない。} \\
$w_{ko}=0.56$ & \ja{私は地球の中心の近くのどこかになっているに違いない。} \\
$w_{ko}=0.58$ & \ja{私は}\ko{지구}\ja{の中心}\ko{의가까이에어딘가에도착하고있어야한다}\ja{。} \\
$w_{ko}=0.60$ & \ko{나는지구의센터의가까이에어딘가에도착하고있어야한다}\ja{。} \\
$w_{ko}=0.70$ & \ko{나는지구의중심근처어딘가에도착해야합니다}\ja{。} \\
$w_{ko}=0.90$ & \ko{나는어딘가지구의중심근처에도착해야합니다}\ja{。} \\
$w_{ko}=1.00$ & \ko{나는어딘가지구의중심근처에도착해야합니다}\ja{。} \\
\rule{0pt}{3ex}
Spanish/Portuguese: & Here the other guinea-pig cheered, and was suppressed. \\
\hline
$w_{pt}=0.00$ & Aquí el otro conejillo de indias animó, y fue suprimido. \\
$w_{pt}=0.30$ & Aquí el otro conejillo de indias animó, y fue suprimido. \\
$w_{pt}=0.40$ & Aquí, o outro porquinho-da-índia alegrou, e foi suprimido. \\
$w_{pt}=0.42$ & Aqui o outro porquinho-da-índia alegrou, e foi suprimido. \\
$w_{pt}=0.70$ & Aqui o outro porquinho-da-índia alegrou, e foi suprimido. \\
$w_{pt}=0.80$ & Aqui a outra cobaia animou, e foi suprimida. \\
$w_{pt}=1.00$ & Aqui a outra cobaia animou, e foi suprimida. \\
\end{tabular}
\end{table}

Table~\ref{table:mixlangexamples} shows examples of mixed target language using
three different multilingual models.  It is interesting that in the first
example (Russian/Belarusian)
the model switches from Russian to Ukrainian (underlined) as target language first
before finally switching to Belarusian. In the second example (Japanese/Korean),
we observe an even more interesting transition from Japanese to Korean, where
the model gradually changes the grammar from Japanese to Korean.
At $w_{ko}=0.58$, the model translates the source
sentence into a mix of Japanese and Korean at the beginning of the target
sentence. At $w_{ko}=0.60$, the source sentence is translated into full Korean,
where all of the source words
are captured, however, the ordering of the words does not look natural.
Interestingly, when the $w_{ko}$ is increased up to $0.7$, the model starts to
translate the source sentence into a Korean sentence that sounds more natural.\footnote{The Korean translation does not contain spaces and uses `\ja{。}'
  as punctuation symbol, and these are all artifacts of applying a Japanese
  postprocessor.}

\section{Conclusion}
We present a simple solution to multilingual NMT. We show
that we can train multilingual NMT models that can be used to translate between
a number of different languages using a single model where all parameters are
shared, which as a positive side-effect also improves the translation
quality of low-resource languages in the mix.
We also show that zero-shot translation without explicit bridging is
possible, which is the first time to our knowledge that a
form of true transfer learning has been shown to work for machine translation.
To explicitly improve the zero-shot translation quality, we explore
two ways of adding available parallel data and find that small
additional amounts are sufficient to reach satisfactory results.
In our largest experiment we merge 12 language pairs into a single model
and achieve only slightly lower translation quality as for the single
language pair baselines despite the drastically
reduced amount of modeling capacity per language in the multilingual model.
Visual interpretation of the results shows that these models learn a form
of interlingua representation between all involved
language pairs. The simple architecture makes it possible to mix
languages on the source or target side to yield some interesting translation
examples. Our approach has been shown to work reliably in a Google-scale
production setting and enables us to scale to a large number of languages
quickly.

\section*{Acknowledgements}
We would like to thank the entire Google Brain Team and Google Translate Team
for their foundational contributions to this project. In particular, we thank
Junyoung Chung for his insights on the topic and Alex Rudnick and Otavio Good
for helpful suggestions.


\bibliography{bnmt_ml}
\bibliographystyle{acm}

\end{document}